\documentclass[conference,compsoc]{IEEEtran}
\ifCLASSOPTIONcompsoc
  \usepackage[nocompress]{cite}
\else
  \usepackage{cite}
\fi

\usepackage{amsmath}

\newtheorem{myDef}{Definition}
\newtheorem{myExample}{Example}
\newtheorem{myProblem}{Problem}
\usepackage[noend]{algpseudocode}

\usepackage{graphicx}
\usepackage{amsmath}
\usepackage{amssymb}
\usepackage{graphicx}
\usepackage{url}
\usepackage{algorithm2e}
\ifCLASSINFOpdf
\else
\fi
\begin{document}
%
\title{Quantization in Layer's Input is Matter}

\author{\IEEEauthorblockN{Cheng Daning}
\IEEEauthorblockA{
Email: chengdn@mail.tsinghua.edu.cn}
\IEEEauthorblockN{Chen Wenguang}
\IEEEauthorblockA{
	Email: cwg@mail.tsinghua.edu.cn}
}
 

%


\maketitle

\begin{abstract}
In this paper, we will show that the quantization in layer's input is more important than parameters' quantization for loss function. And the algorithm which is based on the layer's input quantization error is better than hessian-based mixed precision layout algorithm.
\end{abstract}


%
\IEEEpeerreviewmaketitle

\section{Introduction}
The storage, inference and training of neural networks are highly demanding in terms of both computational resources and time. This is due to the large parameter size of neural networks, which necessitates a large memory footprint and a high number of floating-point operations per second (FLOPS). Thus, it is necessary to develop a compression method for machine learning models. Model quantization, which is based on the strong model computational noise robustness,  is one of the most important means of compressing machine learning models. Computational noise robustness is the index to estimate how effective your algorithm is when noise is introduced during the computing process. The sources of noise are mainly the truncation error and the error in data type conversion in normal cases. 

In model quantization, the high-precision data type initially used for a model's parameters is replaced with a lower-precision data type. For example, it is common to replace FP32 with FP16 in this context, and PyTorch and TensorFlow each include a set of quantization methods that use integers to map to floats. Different quantization methods share the same theoretical basis: using approximate data instead of the original data in the storage and inference processes. A lower-precision data type is less demanding in terms of memory and computational resources, and computations using lower-precision data take less time. In quantization, the source of noise is the precision loss in different quantization level conversion and data type conversion.

The major problem with the model quantization approach is that a naive quantization scheme will likely cause the loss function to increase. It is difficult to replace massive-scale model parameters with ultralow-precision data without significant accuracy loss. In practice, it is impossible to use the same quantization level, i.e., introducing the same level noise for all parameters, for all model parameters while ensuring sufficient performance. 

A means of addressing this problem is to use mixed-precision quantization, in which higher-precision data are used for more ``sensitive'' model parameters and lower-precision data are used for ``nonsensitive'' parameters. In this case, higher-precision data means original data adds small value noise and lower-precision means original data adds large value noise. 

However, the search space for a mixed-precision layout scheme is large. Current algorithms for finding the best mixed-precision layout scheme face the following problems: 1. Most of these algorithms are based on empirical experience and ``fine-tuning'' skill. Some of them avoid neural network and dataset analysis and use only hardware characteristics as the basis for model quantization \cite{2020Effective,2021Pruning}. It is not possible to specify a clear bound to determine under which conditions these algorithms perform better. 2. Some algorithms use Hessian information. Most of them are analysable. However, the computational resource requirements and time cost to obtain Hessian information are also large. 3. Some of these algorithms are beneficial for storage, but to support sufficient inference performance, the quantized model must be saved using high-precision data types in memory.

In this paper, we present a basic analysis of neural networks computational noise robustness. We will show that the model quantization problems for the storage and inference processes require different problem definitions. Moreover, focusing on layerwise post-training static model quantization, we show that in the inference process, the conversion loss due to the different quantization levels at the inputs to different layers plays a dominant role in final accuracy loss. Based on our analysis, we propose three algorithms for addressing different quantization cases.

Our results of analysis are as follows.

1. We define and distinguish between the mathematical descriptions of model quantization for inference and storage.

2. We show that the main factor determining the accuracy loss in the quantization inference process is the influence of different quantization levels conversion in different layers' input. 

3. Based on our analysis, we propose a sef of  layerwise post-training static mixed-precision layout methods in different cases.  And we design experiments to support the theoretical analysis and experimental results show our algorithms are superior to the state-of-the-art (SOTA) analysable-precision layout algorithm HAWQ-v2.

\section{Related Work}
\subsection{Model Compression and Model Quantization}
To reduce the parameter size and computational complexity of machine learning models, model compression methods have undergone great development. Model compression methods include pruning methods, knowledge distillation, weight sharing and quantization methods.

Pruning methods \cite{han2015learning,li2016pruning,mao2017exploring} reduce the nonzero parts of a neural network. These methods are useful for a neural network with large fully connected layers \cite{2019HAWQ}. Knowledge distillation \cite{hinton2015distilling} distils information from a pretrained model. These methods train a neural network to transform a large model into a smaller model. Weight sharing \cite{ullrich2017soft} modifies close parameter values into the same value via clustering methods.

In addition to the above methods, quantization methods \cite{2019HAWQ,morgan1991experimental,courbariaux2015binaryconnect,2020HAWQV3} represent another quantization approach. A common problem definition for quantization is as follows \cite{gholami2021survey}.

\begin{myProblem}
	The objective of quantization is to solve the following optimization problem: \label{ori_pro_def}
	\begin{equation*}
		\min_{q \in \mathbf{Q}}\| q(w) - w\|^2
	\end{equation*}
	where $q$ is the quantization scheme, $q(w)$ is the quantized model with quantization $q$, and $w$ represents the weights, i.e., parameters, in the neural network.
	
\end{myProblem}

Quantization methods replace original data with lower-bit representations. Quantization methods can be divided into post-training quantization and quantization-aware training. Recent studies have given post-training quantization methods an analysable basis that mainly relies on the Hessian matrix. Thus, this paper considers post-training quantization methods. Quantization produces a model with a small memory cost and a high arithmetic intensity without changing the structure of the original neural network. However, quantization methods face the problem that lower precision can lead to significant accuracy degradation.

%

\subsection{Mixed-Precision Layout}
Most quantization methods are designed for mixed-precision quantization \cite{2019HAWQ,2020HAWQV3,dong2019hawq,wu2018mixed,wang2019haq,yu2020search}. In a mixed-precision layout scheme, some layers are stored at higher precision, while others are kept at a lower precision. However, a challenge that must be faced in this approach is how to find the correct mixed-precision settings for the different layers. A brute-force approach is not feasible since the search space is exponentially large in the number of layers.

\subsection{Layerwise Post-training Quantization}
From the perspective of the precision layout, post-training quantization methods can be mainly divided into channelwise \cite{2019Fully,2020Channel}, groupwise \cite{dong2019hawq} and layerwise \cite{2019HAWQ} methods. 

To achieve better accuracy, channelwise quantization and groupwise quantization are widely used. However, these approaches are unfriendly to hardware. In implementations such as cuDNN, different channels are usually combined into a large matrix, and most operations, such as convolution, are transformed into matrix multiplications. When different channels use different data types or precision layouts, such a large matrix must be divided into several pieces, making it impossible to make full use of high-performance Basic Linear Algebra Subprograms (BLAS).

Layerwise mixed-precision layout schemes are more friendly to hardware. Parameters of the same precision are organized together, making full of a program's temporal and spatial locality. In particular, parameters in different layers can be easily addressed by different Internet of Things (IoT) devices, which is beneficial for handling high-volume inference tasks in a pipelined manner.

\section{Background Analysis}
\subsection{Model Computation, Noise Generation and Quantization \label{ch:device cha}}

Compressed models for use in the inference process are computed using different methods depending on the hardware, programming methods and deeplearning framework. All of these methods introduce noise in the computing process.

One reason for this noise problem is that although it is common practice to store and compute model parameters directly using different data types, only data of the same precision can support precise computations in a computer framework. Therefore, before performing computations on nonuniform data, a computer will convert them into the same data type. Usually, in a standard computing environment, a lower-precision data type will be converted into a higher-precision data type; this ensures that the results are correct but requires more computational resources and time. For example, when computing the sum of an integer (int) number and a double-precision (double) number, the C/C++ language will convert the int number into a double number and then compute the sum of two double numbers, which takes more time. However, to accelerate the computing speed, some works on artificial intelligence (AI) computations propose converting higher-precision data types into lower-precision data types based on the premise that AI models are not sensitive to compression noise.

In addition to converting data directly, using a lower-precision data type to linearly map to a higher-precision data type is another important quantization method that is widely used in deep learning frameworks such as PyTorch and TensorFlow. We use the following example to illustrate this method, which is presented in \cite{2020HAWQV3}. Suppose that there are two data objects $input_1$ and $input_2$ are to be subjected to a computing operation, such as multiplication. After the quantization process, we have $Q_1 = \mbox{int}(\frac{input_1}{scale_1})$ and $Q_2 = \mbox{int}(\frac{input_2}{scale_2})$, and we can write
\begin{equation*}
	Q_{output}=\mbox{int}(\frac{input_1 * input_2}{scale_{output}}) \approx\mbox{int}( Q_1Q_2\frac{scale_1*scale_2}{scale_{output}})
\end{equation*}
$scale_{output}$, $scale_1$ and $scale_2$ are precalculated scale factors that depend on the distributions of $input_1$, $input_2$ and the output; $Q_i$ is stored as a lower-precision data type, such as an integer. All $scale$ terms can be precalculated and established ahead of time. Then, throughout the whole inference process, only computations on the $Q_i$ values are needed, which are fast. In this method, the noise is introduced in the $\mbox{int}(\cdot)$ process. This basic idea gives rise to several variants, such as (non)uniform quantization and (non)symmetrical quantization.

\subsection{Computational Noise Robustness and Neural Networks}
In this paper, we mainly use the mathematical properties of extreme points for the analysis of quantization methods. This approach is universal to all cases, not only neural networks. However, there is a myth in the community that it is the neural network properties that guarantee the success of quantization methods \cite{wang2019haq,morgan1991experimental,2020Effective}. To show that instead of the neural network properties, only the properties of extreme points contribute to determining the capability of quantization, i.e., robustness to noise, we must first define the concept of a neural network.

The traditional definition of a neural network \cite{Denilson2016Understanding} as a simulation of the human brain is ambiguous; it is not a rigorous mathematical concept and cannot offer any analysable information. The traditional descriptions of neural networks \cite{Denilson2016Understanding} tend to focus on the inner products of the network weights and inputs, the activation functions and directed acyclic graphs. However, with the development of deep learning, although the majority of neural networks still consist of weighted connections and activation layers, many neural networks no longer obey these rules, such as the network architectures for position embedding and layer norm operations in Transformers. Moreover, current deep learning frameworks, such as PyTorch and TensorFlow, offer application programming interfaces (APIs) to implement any function in a layer. Therefore, we propose that the concept of a neural network follows the engineering concept described by definition \ref{concept nn} rather than a rigorous mathematical definition; in other words, a neural network is an implementation method for a function.

\begin{myDef}
	A function is a neural network if it is implemented in composite function form.
	\label{concept nn}
\end{myDef}

Definition \ref{concept nn} means that a neural network, without training, can be any function. Although with definition \ref{concept nn}, a neural network is no longer a mathematical concept, this idea is widely used in practice \cite{2019Relay}. We can see from definition \ref{concept nn} that the requirement that a neural network be in composite function form is the only mathematical property of a neural network that can be used for analysis. However, it cannot be guaranteed that composite functions are not sensitive to parameters' minor variations and input noise.

\subsection{The Shortcomings of Problem \ref{ori_pro_def}}

Although problem \ref{ori_pro_def} gives researchers a target to aim for when performing quantization, the current problem definition has two shortcomings: 1. The search space of all possible mixed-precision layout schemes is a discrete space that is exponentially large in the number of layers. There is no effective method to solve the corresponding search problem. 2. There is a gap between the problem target and the final task target. As we can see, no terms related to the final task target, such as the loss function or accuracy, appear in the current problem definition. Thus, we need to rebuild the problem description for quantization applications.

However, we find that the targets of the quantization problem vary in different cases. In general, the demand for quantized neural networks arises due to limitations of storage and computation. Thus, we borrow the essential idea of extreme points of the neural network parameters from Eq.~\ref{training problem} and modify the problem definition to address the above two concerns in next chapters.

From the perspective of training, a neural network is the output of an optimization problem whose objective function is
\begin{equation}
	\min_{w}	f(w) =\mathbb{E}_{sample} \ell(w,sample) =\frac{1}{m}\sum_{(x_i,y_i) \in \mathbb{D}} \ell(w,x_i,y_i)
	\label{training problem}
\end{equation}
where $w$ represents the model parameters, $\mathbb{D}$ is the dataset, $m$ is the size of the dataset, $\ell(\cdot)$ is the loss function and $(x_i,y_i)$ represents a sample in the dataset and its label.

As we can see, the well-trained model means that the parameter $w$ is the extreme point of a function. In the following sections, we will further discuss the above definition in the cases of storage and inference and all models, i.e., $w$, are well-trained models in next chapters.

\section{Analysis of Computationsl Noise Robustness and Quantization for Storage}
In the case of quantization for storage, although the quantized model is stored using lower-precision data types, the quantized model is computed at high precision without recovery into the original model value. The following example illustrates this case.

\begin{myExample}
	The original parameters of the model are $(2.1,4.9)$, with (double, double) data types. After quantization, the model parameters are $(2,5)$, which can be stored as (int, int) data. However, when the parameters are used in the computing process, the parameters should be $(2.0,5.0)$, with (double, double) data types.
\end{myExample}

The above case cannot reduce the computational resource costs in the inference and training processes. However, this application can reduce the hard disk cost. In particular, with the development of large AI models, storage is a tightly constrained resource.

The corresponding problem can be defined as follows:

\begin{equation*}
	\hat{f}(w) = f(w+\delta)
\end{equation*}
where $\hat{f}$ is the quantized model and $\delta$ is the noise that is introduced in the quantization process.

Quantization methods for storage solve the following problem:

\begin{equation*}
	\min_{\delta \in \Delta}	\hat{f}(w) - f(w) = f(w+\delta) - f(w) = \frac{\partial f}{\partial w} \delta +\frac{1}{2} \delta^T\frac{\partial^2 f}{\partial w^2} \delta
\end{equation*}
where $\Delta$ is the space of all $\delta$ that satisfy the hardware requirements and the mixed-precision layout requirements.

A well-trained model corresponds to an extreme point of the loss function. Therefore, $\frac{\partial f}{\partial w} = 0$, and
\begin{equation*}
	\min_{\delta \in \Delta}	\hat{f}(w) - f(w) =  \frac{1}{2}\delta^T\frac{\partial^2 f}{\partial w^2} \delta
\end{equation*}

The above descriptions are also presented in the works \cite{2019HAWQ,2020HAWQV3,dong2019hawq,nahshan2021loss}. Accordingly, all these works focus on minimizing $ \delta^T\frac{\partial^2 f}{\partial w^2} \delta$ in a discontinuous space. When the Hessian matrix of $f$ is a block diagonal matrix, HAWQ and HAWQ-v2 are analysable algorithms that can solve this problem.

\section{Analysis of Computational Noise Robustness and Quantization for Inference}
\subsection{Analysis}
\begin{figure}[!htbp]
	\centering
	\includegraphics[width=\columnwidth]{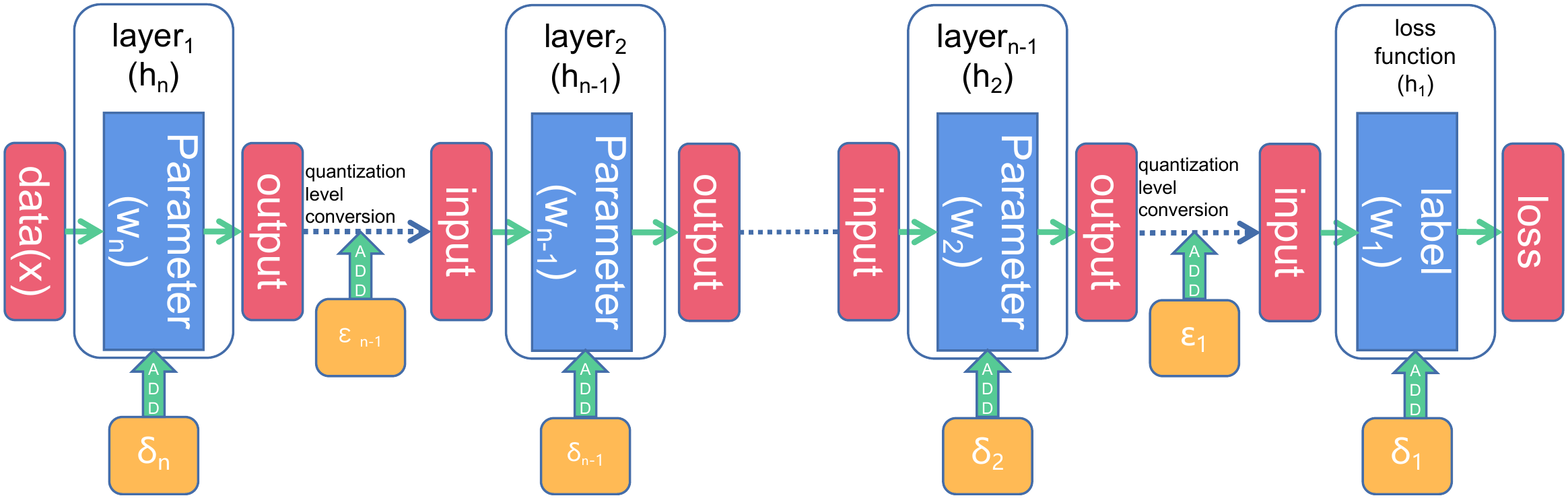}
	\caption{The quantization of models for use in the inference process and the quantization of models for storage are different problems. In storage compression applications, only the parameters are influenced by quantization noise. However, when quantized models are used in the inference process, the outputs of different layers suffer from noise due to the conversion between the different quantization levels of different layers in layerwise quantized models.}
	\label{inference noise}
\end{figure}

Quantization methods for inference are complex. In addition to the noise that is added to the parameters directly, noise is also introduced between different layers in the inference process, as shown in figure \ref{inference noise}, because different quantization levels or data types of different precisions are used in different layers.

Therefore, to provide a clear analysis, we need to divide $\ell(\cdot)$ into subsets of layers based on the fundamental composite function of the neural network. We illustrate the sequential process of neural network analysis, which can also be extended to other cases.

For a sequential ($n-1$)-layer neural network, $\ell(\cdot)$ can be described in the following form.
\begin{small}
\begin{equation*}
	\ell(w,x_i,y_i) = h_1(h_2(h_3(h_4(\cdots h_n(h_{n+1},w_n) \cdots,w_4),w_3),w_2),w_1)
\end{equation*}
\end{small}
where $h_1$ is the loss function, such as the cross-entropy function; $h_i$, $i\in[2,...,n]$, is the ($n-i+1$)th layer in the neural network; $w = (w_n^T,w_{n-1}^T,\cdots,w_1^T)^T$, $w_i$ is the parameter in $h_i(\cdot)$; and for a unified format, $w_1$ is the label $y$ and $h_{n+1}$ is the sample $x$.

After quantization, the quantized loss $\bar{\ell}(\cdot)$ in the inference process is as follows.
\begin{small}
 \begin{align*}
 	&\bar{\ell}(w,x_i,y_i) = \\
 	&h_1(h_2(\cdots h_n(h_{n+1}+\epsilon_n,w_n+\delta_n)+\epsilon_{n-1} \cdots,w_2+\delta_2)+\epsilon_1 ,w_1+\delta_1)
 \end{align*}
\end{small}

where $\delta_i$, $i\in{1,\cdots,n}$, and $\epsilon_i$, $i\in[1,...,n]$, are the minor errors that are introduced in model parameter quantization and in data type conversion in the mixed-precision layout scheme, respectively.

Thus, we obtain the following expression based on the basic total differential calculation.
\begin{align}
	&\bar{\ell}(w,x_i,y_i) - \ell(w,x_i,y_i) \notag \\
	=&\sum_{i=1}^{n} \left( \prod_{j=1}^i \frac{\partial h_j}{\partial h_{j+1}}     \right)  \epsilon_i +\left( \prod_{j=1}^{i-1} \frac{\partial h_j}{\partial h_{j+1}}     \right) \frac{\partial h_i}{\partial w_i} (h_{j+1},w_i) \delta_i \label{get rid of second} \\
	=& \sum_{i=1}^n \frac{\partial \ell}{\partial h_{i+1}} \epsilon_i +  \frac{\partial \ell}{\partial w_i} \delta_i  \label{orignal inference}
\end{align}

We know that for a well-trained model, the expectation of $\ell(\cdot)$'s gradient is zero, i.e., for the $\frac{\partial \ell}{\partial w}$ components, $\frac{\partial \ell}{\partial w_i} = 0$. Therefore, considering the real computational environment, we have the following problem definition for the problem of quantization for inference.
\begin{small}
	 \begin{equation}
	 	\min_{\epsilon \in E }\bar{f}(w) - f(w)= \frac{1}{m}\sum_{(x_j,y_j)\in \mathbb{D}}	  \sum_{i=1}^n \frac{\partial \ell}{\partial h_{i+1}} \epsilon_i =  \frac{1}{m} \sum_{i=1}^n  {\sum_{(x_j,y_j)\in \mathbb{D}}\frac{\partial \ell}{\partial h_{i+1}}} \epsilon_i  
	 	\label{final inference}
	 \end{equation}
\end{small}

where $\bar{f}(w) = \frac{1}{m}\sum \bar{\ell}(\cdot)$. For different quantization conditions, such as using different data types directly or using a lower-precision data type to linearly map to a higher-precision data type as discussed in section \ref{ch:device cha}, $E$ will be slightly different.

As we can see, the problem of quantization for inference is totally different from the problem of quantization for storage. However, current works do not distinguish between the above problems \cite{2019HAWQ,2020HAWQV3,dong2019hawq,nahshan2021loss}; consequently, these works must add a ``fine-tuning'' process, and they still fail in some cases. Moreover, this is the main reason why channelwise quantization methods are successful: in a channel that uses the same data type at all times, the precision loss of the corresponding layer input is usually zero.

\subsection{Application in Post-Training Layerwise Mixed-Precision Quantization}
In practice, one of the issues in model quantization is how to assign different layers to a set of computing devices of fixed but different precision levels. With the continuing development of hardware, various computing devices have emerged that are good at computing on different data types. For example, a typical heterogeneous supercomputing system may consist of some combination of central processing units (CPUs) and graphics processing units (GPUs). CPUs are good at FP64 computing, whereas GPUs can perform computations on the FP32 and FP16 data types quickly. Some custom computing devices, such as tensor processing units (TPUs), may also be designed for certain data types. Given a set of different computing devices, how to assign different layers to these devices is a difficult problem because the layout scheme space is exponentially large in both the number of layers and the number of computing devices.

This issue can be described by a simplified problem in which the different quantization levels for different layers in the whole neural network are fixed in type and quantity, and the objective is to find the mixed-precision layout scheme that minimizes the loss function. We provide the following example to explain this.
\begin{myExample}
	(problem and solution example) We wish to store and compute a five-layer neural network using a set of fixed, mixed precision levels: FP64$*$1, FP32$*$2 and FP16$*$2. We need to determine how to arrange the different layers using the corresponding precision levels so as to minimize the loss given by the loss function.
	
	A possible solution for this example is: The first layer uses FP16 quantization; the second layer uses FP32; The third layer uses FP16 quantization; the fourth layer uses FP64; The fifth layer uses FP32 quantization.
\end{myExample}

For different quantization implementations, the constraint conditions of $\epsilon_i \in E$ are different. For typical cases when a quantization process is used in the PyTorch or TensorFlow framework, the maximum of $\epsilon_i$ is equal to the maximum error associated with the quantization level, i.e., the $scale$ factor in chapter \ref{ch:device cha}. In this case, the problem is described as follows.

\begin{myProblem}
	Find the mixed-precision layout scheme that satisfies
	\begin{align*}
		&\min_{\epsilon \in E }\bar{f}(w) - f(w)= \frac{1}{m}\sum_{(x_j,y_j)\in \mathbb{D}}	  \sum_{i=1}^n \frac{\partial \ell}{\partial h_{i+1}} \epsilon_i \\
		&=  \frac{1}{m} \sum_{i=1}^n  {\sum_{(x_j,y_j)\in \mathbb{D}}\frac{\partial \ell}{\partial h_{i+1}}} \epsilon_i \notag
	\end{align*}
	where $E$ is a space composed of arrangements of fixed precision types and quantities.
	\label{general problem}
\end{myProblem}
Problem \ref{general problem} is solvable because we can obtain an upper bound by using the vector norm.

\begin{align*}
	&\bar{f}(w) - f(w) = \frac{1}{m} \sum_{i=1}^n  {\sum_{(x_j,y_j)\in \mathbb{D}}\frac{\partial \ell}{\partial h_{i+1}}} \epsilon_i \\
	&\le \frac{1}{m} \sum_{i=1}^n  \left\|  {\sum_{(x_j,y_j)\in \mathbb{D}}\frac{\partial \ell}{\partial h_{i+1}}}\right\|  \| \epsilon_i \|
\end{align*}
By Rearrangement Inequality, we can draw the following conclusion: to obtain a lower upper bound, $\| \epsilon_i\|$ should be smaller, i.e., a higher-precision data type or quantization level should be used, when $\left\|  {\sum_{(x_j,y_j)\in \mathbb{D}}\frac{\partial \ell}{\partial h_{i+1}}}\right\| $ is larger. This conclusion leads to algorithm \ref{gerenal algorithm}.

\begin{algorithm}[!htbp]
	\caption{Gradient-based Mixed-Precision Inference Layout Scheme}
	\label{gerenal algorithm}
	
	\KwIn{Neural network $M$, different quantization levels $[q_1,q_2,...,q_n]$}
	\KwOut{Quantized neural network $\bar{M}$}
	
	Arrange $[q_1,q_2,...,q_n]$ in descending order $[q_{i_1},q_{i_2},...,q_{i_n}]$ based on the maximum error $q_i$ introduced;
	
	Compute $\left\|   \sum_{(x_j,y_j)\in \mathbb{D}}  \frac{\partial \ell}{\partial h_{i+1}}\right\| $  for the $i$th layer;
	
	Arrange $[layer_1,layer_2,...,layer_n]$ in ascending order $[layer_{j_1},layer_{j_2},...,layer_{j_n}]$ based on the value of $\left\|  {\sum_{(x_j,y_j)\in \mathbb{D}}\frac{\partial \ell}{\partial h_{i+1}}}\right\| $;
	
	Quantize the $j_k$th layer of the model and the output of the ($j_k - 1$)th layer with $q_{i_k}$ precision;
	
	Combine the quantized layers into $\bar{M}$;
	
	\Return $\bar{M}$
	
\end{algorithm}

However, when the quantization method directly uses different data types to store and compute the model, the constraint condition concerns the fact that when a lower-precision data type is converted into a higher-precision data type, the conversion incurs no loss of precision. Therefore, the problem definition is given as follows in this case.
\begin{myProblem}
	Find the mixed-precision layout scheme that satisfies
	\begin{align*}
		&\min_{\epsilon \in E }\bar{f}(w) - f(w)= \frac{1}{m}\sum_{(x_j,y_j)\in \mathbb{D}}	  \sum_{i=1}^n \frac{\partial \ell}{\partial h_{i+1}} \epsilon_i \\
		& =  \frac{1}{m} \sum_{i=1}^n  {\sum_{(x_j,y_j)\in \mathbb{D}}\frac{\partial \ell}{\partial h_{i+1}}} \epsilon_i \notag\\
		&\mathrm{ s.t. } \epsilon_i =\\ 
		&\begin{cases}
			\epsilon_i& \text{ if $i$th layer is less accurate than ($i-1$)th layer}\\
			0& \text{ if $i$th layer is more accurate than ($i-1$)th layer}
		\end{cases}	\notag
	\end{align*}
	where $E$ is a space composed of arrangements of fixed precision types and quantities.
	\label{inference problem}
\end{myProblem}

The above problem is difficult to solve. However, as an algorithm for solving a similar problem (problem \ref{general problem}), algorithm \ref{gerenal algorithm} can still yield a suboptimal result. Besides algorithm \ref{gerenal algorithm}, the description of problem \ref{inference problem} implies a trivial but important algorithm , i.e., algorithm \ref{trivial algorithm}. Algorithm \ref{trivial algorithm} ensures that $\epsilon_i = 0$ at all times by ensuring that the precision used in the layout is monotonically increasing.

\begin{algorithm}[htbp]
	\caption{Trivial Mixed-Precision Inference Layout Scheme}
	\label{trivial algorithm}
	
	\KwIn{Neural network $M$, different quantization levels $[q_1,q_2,...,q_n]$ where it is lossless when low precision quantization levels converse to high precision quantization level }
	\KwOut{Quantized neural network $\bar{M}$}
	
	Arrange $[q_1,q_2,...,q_n]$ in descending order $[q_{i_1},q_{i_2},...,q_{i_n}]$ based on the maximum error $q_i$ introduced;
	
	Quantize the $j$th layer of the model and the output of the ($j-1$)th layer with $q_{i_j}$ precision;
	
	Combine the quantized layers into $\bar{M}$;
	
	\Return $\bar{M}$;
	
\end{algorithm}

Algorithm \ref{trivial algorithm} is trivial, but it is important because in many cases, such as a fully connected neural network, the number of parameters per layer decreases closer to the output. In these cases, it is natural that algorithm \ref{trivial algorithm} will solve problem \ref{inference problem}.  We present example \ref{trival algorithm example} to explain this algorithm.

\begin{myExample}
	We wish to store and compute a three-layer fully connected neural network using a layout with fixed, mixed precision levels of FP64$*$1, FP32$*$1 and FP16$*$1. The best layout scheme that costs the least memory and supports the fastest inference is the scheme in which FP16 is used for storage and computation of the input data and in the first layer, FP32 is used for storage and computation of the output from the first layer and in the second layer, and FP64 is used for storage and computation of the output from the second layer and in the last layer.
	\label{trival algorithm example}
\end{myExample}

In practice, this layout method is found by numerous experiments and used in many applications \cite{qingdaochaosuan}, but they cannot explain why this layout is reasonable.

\subsection{Combination of Inference and Storage}

The problem definitions for inference and storage conflict; however, the tasks of inference and storage are compatible because they occur in different phases. It is possible to load a model quantized for storage into a full-precision model and then quantize this model for inference.

\begin{algorithm}[!ht]
	\caption{Gradient-based Mixed-Precision Inference Layout Scheme after Storage Quantization}
	\label{parameter in count algorithm}
	
	\KwIn{Neural network $M$ quantized for storage, different quantization levels $[q_1,q_2,...,q_n]$}
	\KwOut{Quantized neural network $\bar{M}$}
	
	Arrange $[q_1,q_2,...,q_n]$ in descending order $[q_{i_1},q_{i_2},...,q_{i_n}]$ based on the maximum error $q_i$ introduced;
	
	Compute $\left\|   \sum_{(x_j,y_j)\in \mathbb{D}}  \frac{\partial \ell}{\partial h_{i+1}}\right\| $  and  $\left\|   \sum_{(x_j,y_j)\in \mathbb{D}}  \frac{\partial \ell}{\partial w_{i}}\right\| $ for the $i$th layer;
	
	\If{the layers' parameters and input share the same quantization level}{
		Arrange $[layer_1,layer_2,...,layer_n]$ in ascending order $[layer_{j_1},layer_{j_2},...,layer_{j_n}]$ based on the value of   $\left\|  {\sum_{(x_j,y_j)\in \mathbb{D}}\frac{\partial \ell}{\partial h_{i+1}}}\right\| $ +   $\left\|   \sum_{(x_j,y_j)\in \mathbb{D}}  \frac{\partial \ell}{\partial w_{i}}\right\| $;
		
		Quantize the model parameters in the $j_k$th layer and the output of the ($j_k - 1$)th layer with $q_{i_k}$ precision;
	}\Else{
		
		Arrange $[layer_1,layer_2,...,layer_n]$ in ascending order $[layer_{j_1},layer_{j_2},...,layer_{j_n}]$ based on the value of $\left\|  {\sum_{(x_j,y_j)\in \mathbb{D}}\frac{\partial \ell}{\partial h_{i+1}}}\right\| $;
		
		Arrange $[layer_1,layer_2,...,layer_n]$ in ascending order $[layer_{v_1},layer_{v_2},...,layer_{v_n}]$ based on the value of    $\left\|   \sum_{(x_j,y_j)\in \mathbb{D}}  \frac{\partial \ell}{\partial w_{i}}\right\| $;
		
		Quantize the model parameters in the $v_k$th layer with $q_{i_k}$ precision and the output of the ($j_k - 1$)th layer with $q_{i_k}$ precision;
	}
	
	Combine the quantized layers into $\bar{M}$;
	
	\Return $\bar{M}$
	
\end{algorithm}

The above approach faces the problem that the noise introduced in the storage quantization process will change the extreme point properties because the parameters' gradient may be large: Some neural networks, the maximum eigenvalue of the Hessian matrix is large, and the minor change of parameters may lead to the signifiant change of parameters' gradient. So the analysis from Eq. \ref{orignal inference} to Eq. \ref{final inference} fails.

To deal with above problem, we need to modify algorithm \ref{gerenal algorithm} into algorithm \ref{parameter in count algorithm} from Eq. \ref{orignal inference}: As seen from Eq. \ref{orignal inference}, quantization for inference does not require a model that is well trained. Therefore, we can treat the model quantized for storage as an unconverged model to take $\left\|  {\sum_{(x_j,y_j)\in \mathbb{D}}\frac{\partial \ell}{\partial w_i}} \right\| $ into account where the $w_i$ is the parameters in quantized model in storage, and assign the quantization level to parameters based on  $\left\|  {\sum_{(x_j,y_j)\in \mathbb{D}}\frac{\partial \ell}{\partial w_i}} \right\| $.

\section{Experiment}
\subsection{Experimental Setting}
In this section, we match the theoretical analysis to real neural network applications. We compare algorithm \ref{gerenal algorithm} and HAWQ-v2 \cite{dong2019hawq}, which is a SOTA quantization layout scheme algorithm and has a clear analysis process that can guarantee its performance and scope of application. HAWQ-v2 also is used in HAWQ-v3 \cite{2020HAWQV3} as the mixed precision layout scheme algorithm. Because ResNets do not fit the case in which the number of neural network parameters decreases in layers closer to the output and they share the same analysis, we do not consider algorithm \ref{trivial algorithm} in this section. To show that the precision loss in the quantization processes for different layers is the main factor affecting the loss function, we do consider algorithm \ref{parameter in count algorithm} for comparison. Algorithm \ref{parameter in count algorithm} also presents the effect of combining inference and storage quantization.

In these experiments, we use the CIFAR-10 dataset and ResNets. Considering the memory cost and time cost for computing the Hessian and its trace, we use ResNet-8 and ResNet-14.


To avoid discussing the effect of different quantization skills, like (non)uniform quantization, (non)symmetrical quantization, we use a stronger and more general experiments method: adding random noise, $\sigma$, to the corresponding parts of the models with different level. This experiments method also directly matches our analysis in paper. In these experiments, we use four levels of noise to add to the inputs to the different layers and the layer parameters, as described below.

$\sigma_1$ is a random number in the range $[-1*10^{-3},1*10^{-3}]$. We choose $-1*10^{-3}$ as our maximum noise because in practice, the minimal positive number for FP16 is between $6.104×10^(-5)$ which is the precision loss, and we enlarge it into $1*10^{-3}$. FP16 is the most common data type chosen for the storage and computation of quantized models. $\sigma_4 $ is a random number in the range $ [-1*10^{-10},1*10^{-10}]$. We choose $-1*10^{-10}$ as our minimum noise because in our preliminary tests, when the precision loss was below $1*10^{-10}$, the change in the loss function for ResNet on CIFAR-10 was below $1*10^{-5}$, which is not significant. To consider various levels of precision loss, we also choose $\sigma_2 \in [-1*10^{-5},1*10^{-5}]$ and $\sigma_3 \in [-1*10^{-7},1*10^{-7}]$ as intermediate cases between $\sigma_1$ and $\sigma_4$.

In the experiments with HAWQ-v2 and algorithm \ref{gerenal algorithm}, for each model, we divide their layers into four subsets based on HAWQ-v2 and algorithm \ref{gerenal algorithm}, and we add one kind of noise to the parameters and input for each subset. To match our analysis, we do not distinguish the weights or activation functions in these experiments because we believe that all kinds of layers play the same role in our method. In the HAWQ-v2 experiments, we set the quantization levels based on the mathematical expectation of the traces of the Hessian matrices of the different layers' parameters. In the experiments with algorithm \ref{gerenal algorithm}, we set the quantization levels based on the expectation of the norms of the layers' input gradients.

For the experiments with algorithm \ref{parameter in count algorithm}, we also divide the layers into four subsets, and we add $\sigma_1$ to all layers' parameters to simulate the recovered quantized model. The choice of quantization for the parameters and the layer inputs depends on the sum of the input gradient and the parameter gradient. When adding noise, for the parameter layer to which $\sigma_1$ should be added, 0 is added instead because in practice, the recovery and requantization of a quantized model at the same quantization level is a lossless process.

In all experiments, we choose 256 samples from the dataset and calculate the corresponding values used in the layout arrangement.
\subsection{Experimental Results and Analyses}

In the ResNet-8 experiments, the value of the loss function at full precision is 0.3896, the value of the loss function using HAWQ-v2 is 0.42678, the value of the loss function using algorithm \ref{gerenal algorithm} is 0.3971, and the value of the loss function using algorithm \ref{parameter in count algorithm} is 0.3980. In the ResNet-14 experiments, the value of the loss function at full precision is 0.3161, the value of the loss function using HAWQ-v2 is 0.3284, the value of the loss function using algorithm \ref{gerenal algorithm} is 0.3179, and the value of the loss function using algorithm \ref{parameter in count algorithm} is 0.3210.

The above experimental results show that the precision loss in the inputs to the different layers exerts more influence on the final loss function than the precision loss in the model parameters: 1. Algorithm \ref{gerenal algorithm} is better than HAWQ-v2 in terms of the loss function, computational resource consumption and time cost. 2. In the experiments with algorithm \ref{parameter in count algorithm}, although the parameter gradients are nonzero, they are still small. The layout scheme is the same as that in the experiments with algorithm \ref{gerenal algorithm}. Compared with the algorithm \ref{gerenal algorithm} experiments, the only difference in the algorithm \ref{parameter in count algorithm} experiments is that the noise in the parameters is larger.

\section{Conclusion}
In this paper, we divide the model quantization problem into the problems of quantization for inference and quantization for storage. In the inference quantization problem, we show that the precision loss in the inputs to different layers plays a dominant role, both theoretically and in experiments. Based on this property, we design three algorithms that are superior to HAWQ-v2 in terms of performance and resource consumption.

\bibliographystyle{IEEEtran}
\bibliography{mybibtex.bib}

\begin{thebibliography}{10}
\providecommand{\url}[1]{#1}
\csname url@samestyle\endcsname
\providecommand{\newblock}{\relax}
\providecommand{\bibinfo}[2]{#2}
\providecommand{\BIBentrySTDinterwordspacing}{\spaceskip=0pt\relax}
\providecommand{\BIBentryALTinterwordstretchfactor}{4}
\providecommand{\BIBentryALTinterwordspacing}{\spaceskip=\fontdimen2\font plus
\BIBentryALTinterwordstretchfactor\fontdimen3\font minus
  \fontdimen4\font\relax}
\providecommand{\BIBforeignlanguage}[2]{{%
\expandafter\ifx\csname l@#1\endcsname\relax
\typeout{** WARNING: IEEEtran.bst: No hyphenation pattern has been}%
\typeout{** loaded for the language `#1'. Using the pattern for}%
\typeout{** the default language instead.}%
\else
\language=\csname l@#1\endcsname
\fi
#2}}
\providecommand{\BIBdecl}{\relax}
\BIBdecl

\bibitem{2020Effective}
A.~Demidovskij and E.~Smirnov, ``Effective post-training quantization of neural
  networks for inference on low power neural accelerator,'' in \emph{2020
  International Joint Conference on Neural Networks (IJCNN)}, 2020.

\bibitem{2021Pruning}
T.~Liang, J.~Glossner, L.~Wang, and S.~Shi, ``Pruning and quantization for deep
  neural network acceleration: A survey,'' 2021.

\bibitem{han2015learning}
S.~Han, J.~Pool, J.~Tran, and W.~J. Dally, ``Learning both weights and
  connections for efficient neural networks,'' \emph{arXiv preprint
  arXiv:1506.02626}, 2015.

\bibitem{li2016pruning}
H.~Li, A.~Kadav, I.~Durdanovic, H.~Samet, and H.~P. Graf, ``Pruning filters for
  efficient convnets,'' \emph{arXiv preprint arXiv:1608.08710}, 2016.

\bibitem{mao2017exploring}
H.~Mao, S.~Han, J.~Pool, W.~Li, X.~Liu, Y.~Wang, and W.~J. Dally, ``Exploring
  the regularity of sparse structure in convolutional neural networks,''
  \emph{arXiv preprint arXiv:1705.08922}, 2017.

\bibitem{2019HAWQ}
Z.~Dong, Z.~Yao, A.~Gholami, M.~Mahoney, and K.~Keutzer, ``Hawq: Hessian aware
  quantization of neural networks with mixed-precision,'' \emph{IEEE}, 2019.

\bibitem{hinton2015distilling}
G.~Hinton, O.~Vinyals, and J.~Dean, ``Distilling the knowledge in a neural
  network,'' \emph{arXiv preprint arXiv:1503.02531}, 2015.

\bibitem{ullrich2017soft}
K.~Ullrich, E.~Meeds, and M.~Welling, ``Soft weight-sharing for neural network
  compression,'' \emph{arXiv preprint arXiv:1702.04008}, 2017.

\bibitem{morgan1991experimental}
N.~Morgan \emph{et~al.}, ``Experimental determination of precision requirements
  for back-propagation training of artificial neural networks,'' in \emph{Proc.
  Second Int’l. Conf. Microelectronics for Neural Networks}.\hskip 1em plus
  0.5em minus 0.4em\relax Citeseer, 1991, pp. 9--16.

\bibitem{courbariaux2015binaryconnect}
M.~Courbariaux, Y.~Bengio, and J.-P. David, ``Binaryconnect: Training deep
  neural networks with binary weights during propagations,'' in \emph{Advances
  in neural information processing systems}, 2015, pp. 3123--3131.

\bibitem{2020HAWQV3}
Z.~Yao, Z.~Dong, Z.~Zheng, A.~Gholami, and K.~Keutzer, ``Hawqv3: Dyadic neural
  network quantization,'' 2020.

\bibitem{gholami2021survey}
A.~Gholami, S.~Kim, Z.~Dong, Z.~Yao, M.~W. Mahoney, and K.~Keutzer, ``A survey
  of quantization methods for efficient neural network inference,'' \emph{arXiv
  preprint arXiv:2103.13630}, 2021.

\bibitem{dong2019hawq}
Z.~Dong, Z.~Yao, Y.~Cai, D.~Arfeen, A.~Gholami, M.~W. Mahoney, and K.~Keutzer,
  ``Hawq-v2: Hessian aware trace-weighted quantization of neural networks,''
  \emph{arXiv preprint arXiv:1911.03852}, 2019.

\bibitem{wu2018mixed}
B.~Wu, Y.~Wang, P.~Zhang, Y.~Tian, P.~Vajda, and K.~Keutzer, ``Mixed precision
  quantization of convnets via differentiable neural architecture search,''
  \emph{arXiv preprint arXiv:1812.00090}, 2018.

\bibitem{wang2019haq}
K.~Wang, Z.~Liu, Y.~Lin, J.~Lin, and S.~Han, ``Haq: Hardware-aware automated
  quantization with mixed precision,'' in \emph{Proceedings of the IEEE/CVF
  Conference on Computer Vision and Pattern Recognition}, 2019, pp. 8612--8620.

\bibitem{yu2020search}
H.~Yu, Q.~Han, J.~Li, J.~Shi, G.~Cheng, and B.~Fan, ``Search what you want:
  Barrier panelty nas for mixed precision quantization,'' in \emph{European
  Conference on Computer Vision}.\hskip 1em plus 0.5em minus 0.4em\relax
  Springer, 2020, pp. 1--16.

\bibitem{2019Fully}
R.~Li, Y.~Wang, F.~Liang, H.~Qin, and R.~Fan, ``Fully quantized network for
  object detection,'' in \emph{2019 IEEE/CVF Conference on Computer Vision and
  Pattern Recognition (CVPR)}, 2019.

\bibitem{2020Channel}
X.~Qian, V.~Li, and C.~Darren, ``Channel-wise hessian aware trace-weighted
  quantization of neural networks,'' 2020.

\bibitem{Denilson2016Understanding}
Denilson and Barbosa, ``Understanding machine learning: from theory to
  algorithms,'' \emph{Computing reviews}, vol.~57, no.~4, pp. 238--238, 2016.

\bibitem{2019Relay}
J.~Roesch, S.~Lyubomirsky, M.~Kirisame, J.~Pollock, and Z.~Tatlock, ``Relay: A
  high-level ir for deep learning,'' 2019.

\bibitem{nahshan2021loss}
Y.~Nahshan, B.~Chmiel, C.~Baskin, E.~Zheltonozhskii, R.~Banner, A.~M.
  Bronstein, and A.~Mendelson, ``Loss aware post-training quantization,''
  \emph{Machine Learning}, vol. 110, no.~11, pp. 3245--3262, 2021.

\bibitem{qingdaochaosuan}
M.~Zixuan, H.~Jiaao, Q.~Jiezhong, C.~Huanqi, and W.~Yuanwei, ``Bagualu:
  Targeting brain scale pretrained models with over 37 million cores,'' in
  \emph{ACM SIGPLAN Annual Symposium Principles and Practice of Parallel
  Programming}.\hskip 1em plus 0.5em minus 0.4em\relax ACM, 2022.

\end{thebibliography}

\end{document}